\documentclass[preprint,5p,times,twocolumn]{elsarticle}

\usepackage{amssymb}
\usepackage{amsmath}
\usepackage{multirow}
\usepackage{booktabs}
\usepackage{comment}
\usepackage{adjustbox}
\usepackage{float}
\usepackage{fancyhdr}
\usepackage{url}
\usepackage{hyperref}
\usepackage[compact]{titlesec}
\usepackage[
 font=footnotesize 
 ]{subfig}
\usepackage{lineno}
\usepackage{graphicx}
\usepackage{xcolor}
\usepackage{hyperref}
\usepackage{amsmath}
\usepackage{caption}
\usepackage{colortbl}

\journal{Elsevier.}

\titlespacing{\section}{0.2pt}{0.4ex}{0.2ex}
\titlespacing{\subsection}{0.2pt}{0.4ex}{0.2ex}
\titlespacing{\subsubsection}{0pt}{0.2ex}{0.2ex}

\begin{document}

\begin{frontmatter}

\title{DG-DETR: Toward Domain Generalized Detection Transformer}

\author[a]{Seongmin Hwang}
\ead{sm.hwang@gm.gist.ac.kr}

\author[b]{Daeyoung Han}
\ead{xesta120@gist.ac.kr}

\author[b]{Moongu Jeon\corref{cor1}}
\ead{mgjeon@gist.ac.kr}

\cortext[cor1]{Corresponding author.}

\affiliation[a]{organization={Artificial Intelligence Graduate School, Gwangju Institute of Science and Technology (GIST)}, 
            city={Gwangju},
            postcode={61005}, 
            country={Republic of Korea}
            }

\affiliation[b]{organization={School of Electrical Engineering and Computer Science, Gwangju Institute of Science and Technology (GIST)}, 
            city={Gwangju},
            postcode={61005}, 
            country={Republic of Korea}
            }
        
\begin{abstract}
End-to-end Transformer-based detectors (DETRs) have demonstrated strong detection performance. However, domain generalization (DG) research has primarily focused on convolutional neural network (CNN)-based detectors, while paying little attention to enhancing the robustness of DETRs. In this letter, we introduce a Domain Generalized DEtection TRansformer (DG-DETR), a simple, effective, and plug-and-play method that improves out-of-distribution (OOD) robustness for DETRs. Specifically, we propose a novel domain-agnostic query selection strategy that removes domain-induced biases from object queries via orthogonal projection onto the instance-specific style space. Additionally, we leverage a wavelet decomposition to disentangle features into domain-invariant and domain-specific components, enabling synthesis of diverse latent styles while preserving the semantic features of objects. Experimental results validate the effectiveness of DG-DETR. Our code is available at \href{https://github.com/smin-hwang/DG-DETR}{https://github.com/smin-hwang/DG-DETR}.
\end{abstract}

\begin{keyword}
Object detection\sep
Domain generalization\sep
Detection transformer\sep
Query selection\sep
Wavelet decomposition.
\end{keyword}

\end{frontmatter}

\section{Introduction}
\label{sec:intro}
Deep neural network (DNN)-based object detectors~\cite{fasterrcnn, redmon2016you, fcos, DETR} have shown remarkable performance in various computer vision tasks under the independent and identically distributed (\textit{i.i.d.}) assumption that training and test data are sampled from the same distribution. However, when these models are deployed in real-world scenarios where the distribution differs from the training set, their performance often degrades significantly due to domain-shifts \cite{ben2010theory, chen2018domain, sakaridis2018semantic}. This issue is especially critical in safety-sensitive applications such as autonomous driving. In this regard, addressing the distribution shifts between source and target domains has remained a long-standing challenge in the computer vision community.

To alleviate this problem, the prominent research direction is unsupervised domain adaptation (UDA) \cite{inoue2018cross, deng2021unbiased, li2022cross, cao2023contrastive}, which aims to align the data distribution of a labeled source domain to that of an unlabeled target domain. Although UDA methods have demonstrated promising results, they heavily rely on the assumption that target domain data is available during the adaptation process. Moreover, collecting sufficient training data for all target domains of interest is often impractical \cite{shu2021open} due to the cost and effort required, even without annotations.

Domain generalization (DG) \cite{balaji2018metareg, dou2019domain, li2018domain, shu2021open} has emerged as a feasible solution, aiming to train a model that generalizes well to unseen target domains by learning from multiple observed source domains during training. Most DG methods generally strive to learn domain-invariant feature representations across these source domains, but their performance is sensitive to the diversity and quantity of the available source domains \cite{choi2021robustnet, zhou2022domain}, which are often costly and labor-intensive to collect. As a more practical approach, recent research has begun focusing on single-domain generalization (S-DG) \cite{wang2021learning, fan2021adversarially}, which aims to train robust models using only one source domain. However, achieving out-of-distribution (OOD) robustness in this setting remains an open challenge.

In recent years, research on S-DG for object detection \cite{wu2022single, fan2023towards, lee2024object} has been limited, despite its critical importance in many visual perceptual systems. Furthermore, existing S-DG studies have largely focused on CNN-based detectors \cite{fasterrcnn, redmon2016you, fcos}, although Vision Transformers (ViTs) have demonstrated strong generalization capabilities attributed to their superior shape bias, which stems from their ability to capture global dependencies among tokens \cite{bai2022improving, aloft}. Building on this, in this letter, we explore the potential of Detection Transformer (DETR) \cite{DETR}, which shares the global modeling advantage, for single-domain generalization in object detection.

 \begin{figure*}[ht]
 \centering
    \includegraphics[scale=0.50]{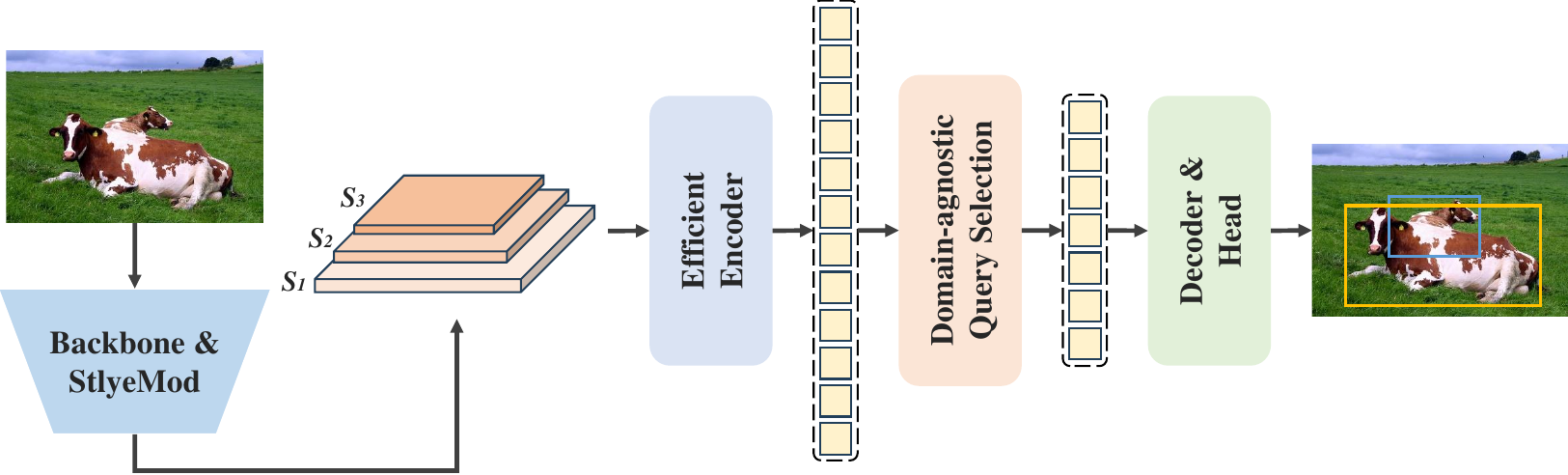}
    \caption{Overview of the proposed DG-DETR framework. The backbone extracts multi-scale features, represented as $s_1$, $s_2$, and $s_3$, from its last three stages. These features are processed by the encoder and fed into the Domain-agnostic Query Selection (DAQS) module. The DAQS module explicitly removes domain-induced biases from object queries, providing high-quality initial queries for the decoder.
    }
    \label{fig1:overview}
    \vspace{0.2cm}
 \end{figure*}

Similar to the human visual system, which exhibits high robustness by extracting global shape information \cite{geirhosimagenet}, recent studies \cite{shi2020informative, naseer2021intriguing, zhang2022delving} have shown that shape-biased models achieve high robustness. Inspired by this observation, \textit{we aim to increase shape bias both implicitly and explicitly to improve model robustness against domain shifts.} In this work, we propose a domain-agnostic query selection strategy to explicitly remove domain-induced biases from queries and select those that contain rich domain-invariant representations (\textit{e.g.}, shape). This approach provides high-quality initial queries to the decoder for robust object detection.

To implicitly improve model robustness, style augmentation \cite{liuncertainty, zhoudomain}, which assumes that domain shifts are caused by style variance, has been successfully applied to single-domain generalization in image classification, generating multiple domains from a single source domain. However, style synthesis may potentially distort image contents, which are essential for the hierarchical object detection where a diverse context is present. Therefore, the key challenge in applying style augmentation to object detection lies in \textit{preserving the shape and semantics of objects while achieving diverse style augmentation}. To do this, we utilize wavelet transform to decompose the domain-invariant features and domain-specific features. This approach enables us to perturb only the domain-specific features (\textit{e.g.}, texture), while preserving the content structures (\textit{e.g.}, shape) inherent in the images.
The key contributions of this letter are summarized as follows:
\begin{itemize}
    \item To the best of our knowledge, this is the first study to explore the potential of DETRs for domain generalized object detection and to propose DG-DETR as a simple yet effective plug-and-play solution.
    
    \item We propose a domain-agnostic query selection strategy that explicitly removes domain-induced biases from object queries.

    \item To preserve the semantics of objects during style augmentation, we leverage wavelet decomposition to perturb only domain-specific features.
\end{itemize}

\section{Related Work}
\label{sec:related_work}
\subsection{Single-Domain Generalization for Object Detection}
To improve model robustness in unseen target domains, the single-domain generalization (S-DG) setting is practical as it addresses the common issue of data scarcity from multiple domains. Recent advances in S-DG can be categorized into four main approaches: data augmentation~\cite{lee2024object, danish2024improving}, feature augmentation~\cite{liuncertainty, zhoudomain, vidit2023clip}, domain-invariant feature learning~\cite{wang2021robust, wu2022single}, and architectural improvements~\cite{wu2024g}. The most common approach is data augmentation, which aims to diversify the distribution of source data. Lee et al.~\cite{lee2024object} use object-aware augmentation and contrastive loss to improve model robustness. DivAlign~\cite{danish2024improving} employs carefully selected augmentations to diversify the source domain and further aligns detections across multiple augmented views based on classification and localization. In addition to data augmentation, feature augmentation methods improve generalization by perturbing feature statistics~\cite{liuncertainty}, mixing novel styles~\cite{zhoudomain}, or leveraging vision-language models for semantic augmentation~\cite{vidit2023clip}. Other approaches aim to learn domain-invariant representations or to improve network architectures for enhanced generalization. Wu et al.~\cite{wu2022single} proposed a cyclic-disentangled self-distillation (CDSD) to extract domain-invariant representations, while G-NAS~\cite{wu2024g} uses neural architecture search to find more generalizable model structures. While these approaches have advanced single-domain generalization for CNN-based detectors, they are not directly tailored to transformer-based detectors. In this work, we address this gap by introducing DG-DETR with domain-agnostic query selection and wavelet-based feature perturbation to improve OOD robustness in the single-domain generalization setting.
\vspace{0.2cm}

\subsection{End-to-end Object Detectors}
Transformer-based detectors have recently emerged as a powerful alternative to CNN-based detectors by directly formulating object detection as a set prediction problem. Carion et al.~\cite{DETR} first propose the transformer-based end-to-end detector, DEtection TRansformer (DETR), which eliminates hand-crafted components but suffers from slow convergence. To address this, Deformable-DETR~\cite{zhudeformable} introduces multi-scale deformable attention, significantly improving training efficiency and detection accuracy. Subsequent advances further refine query design, such as dynamic anchor boxes queries~\cite{liu2022dab}, denoising queries~\cite{li2022dn, zhangdino}, and adopt hybrid matching strategies~\cite{jia2023detrs}. For deployment efficiency, RT-DETR~\cite{zhao2024detrs} further demonstrates that real-time and end-to-end detection can be achieved through an efficient hybrid encoder and an IoU-aware query selection strategy.
\vspace{0.2cm}

\section{Proposed Method}
\label{sec:method}

In this section, we briefly introduce the overall architecture of DG-DETR. We then describe the newly introduced components for robust object detection in detail. 
\vspace{0.2cm}

\subsection{Overview}
DETRs consist of a CNN backbone, an encoder-decoder structure, and object class and box position predictors. To improve generalization capability, we introduce a style augmentation module and domain-agnostic query selection. The overall framework of DG-DETR is illustrated in Fig. \ref{fig1:overview}.

Given an input image, the backbone network first extracts image features, while the style augmentation module synthesizes diverse latent styles. The output is then fed into an efficient hybrid encoder~\cite{zhao2024detrs} which combines Transformer and CNN to enhance feature representation. The encoded features are subsequently passed to the domain-agnostic query selection module. This module removes domain-induced biases from queries and provides high-quality initial queries for the decoder. Finally, the decoder with auxiliary prediction heads iteratively refines object queries to generate object categories and bounding boxes. We note that while RT-DETR~\cite{zhao2024detrs} is adopted as an example, the DG-DETR framework is compatible with other DETR-based detectors.
\vspace{0.2cm}

 \begin{figure}[t]
 \centering
    \includegraphics[scale=0.26]{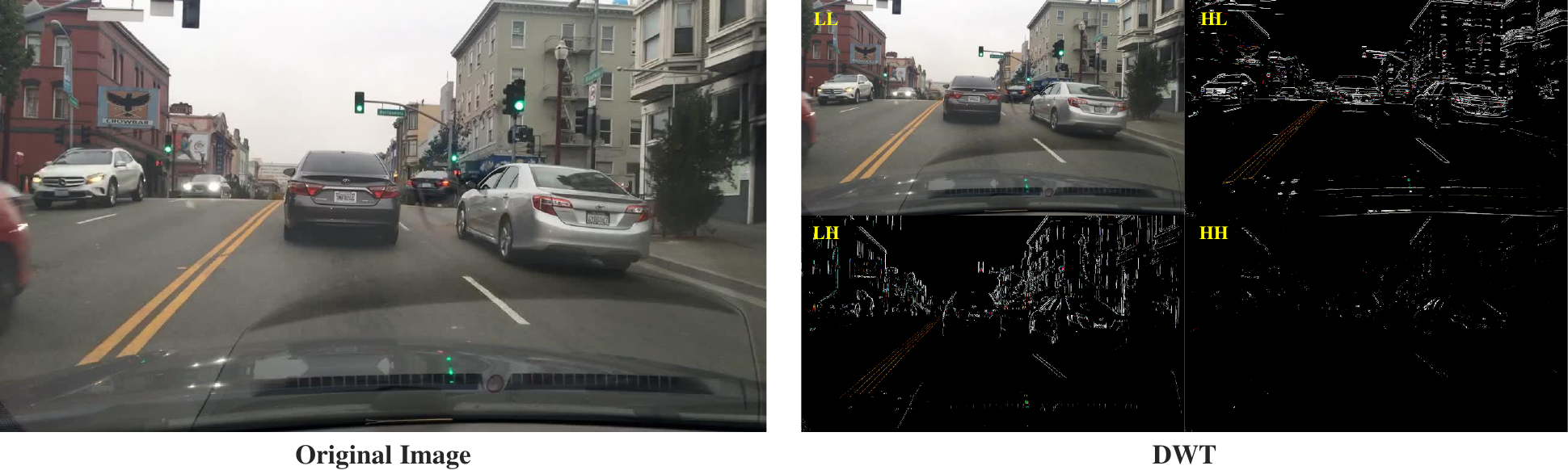}
    \caption{Visualization of wavelet decomposition. The image is decomposed into low- and high-frequency components. In the Discrete Wavelet Transform (DWT) result, LL represents the low-frequency approximation, capturing the overall style. LH, HL, and HH represent the high-frequency detail coefficients (horizontal, vertical, and diagonal details, respectively), which primarily preserve object shapes and edge information.
    }
    \label{fig:dwt}
    \vspace{0.2cm}
 \end{figure}

 \begin{figure*}[t]
 \centering
    \includegraphics[scale=0.48]{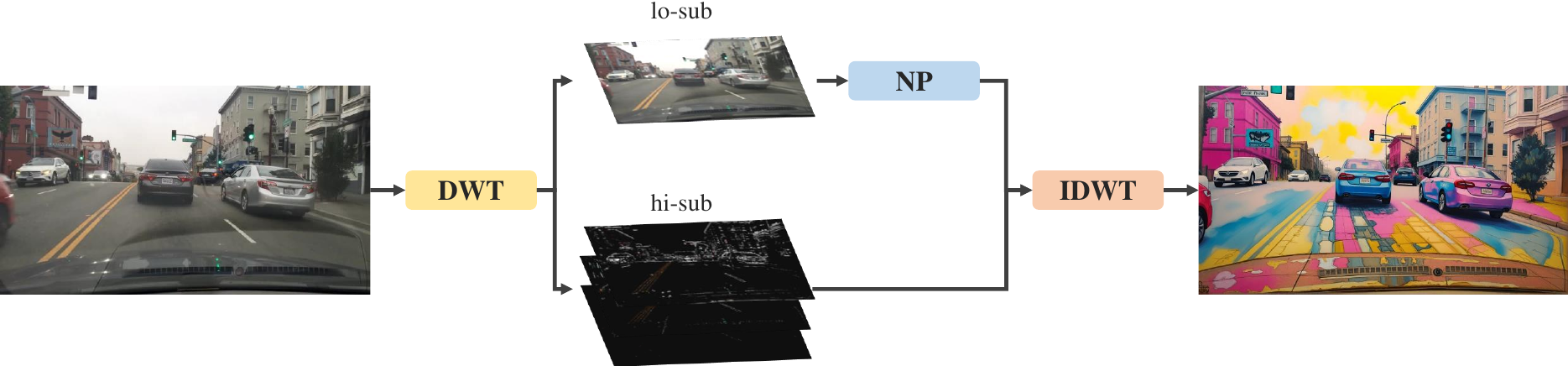}
    \caption{Overview of our style augmentation module. Wavelet transform decomposes the shallow CNN features into low- and high-frequency components. We perturb only the low-frequency features components.}
    \label{fig2:wavenp}
    \vspace{0.2cm}
 \end{figure*}

\subsection{Wavelet-guided Style Augmentation Module}
Augmenting styles potentially cause the problem of semantic drift in objects, leading to imprecise object localization and misclassification. Inspired by the properties of the frequency spectrum \cite{yoo2019photorealistic, aloft}, we argue that frequency-based style augmentation enables the synthesis of diverse domains while preserving object semantics. Recall that the high-frequency components capture more global features (\textit{e.g.}, shape) while the low-frequency components contain smooth surfaces and textures (see Fig. \ref{fig:dwt}). This suggests that applying style augmentation only to the low-frequency components affects overall texture while preserving the image contents.

To this end, we propose to utilize wavelet transform to disentangle high- and low-frequency features of images. Specifically, given an intermediate CNN feature map $F \in \mathbb{R}^{H \times W \times C}$ with spatial dimensions $H \times W$ and $C$ channels, discrete wavelet transform (DWT) applies four kernels, $LL^T, LH^T, HL^T$, and $HH^T$, where the low-pass filter is defined as $L=\left[\begin{array}{ll}1 / \sqrt{2} & 1 / \sqrt{2}\end{array}\right]$ and high pass filter is $H=\left[\begin{array}{ll}-1 / \sqrt{2} & 1 / \sqrt{2}\end{array}\right]$. These kernels slide across the input $F$ to decompose it into the four wavelet sub-bands: $F_{ll}, F_{lh}, F_{hl},$ and $F_{hh}$, each with reduced resolution. In this letter, we denote the low-frequency components $F_{ll}$ as $F_{low}$ and the high-frequency components $[F_{lh}, F_{hl}, F_{hh}]$ as $F_{high}$, respectively.

As shallow CNN layers preserve more style information \cite{pan2018two}, we apply the style augmentation module to the CNN backbone. Specifically, as a common practice, we adopt the feature channel statistics (\textit{i.e.}, mean and variance) to represent image styles. We then perturb the style statistics of source domain training instances to synthesize novel domain styles. Several prior studies~\cite{zhoudomain, liuncertainty, jeon2021feature, fan2023towards} have explored domain synthesis in feature space by perturbing feature statistics. To balance diversity and fidelity, we employ Normalization Perturbation (NP)~\cite{fan2023towards} to perturb feature statistics in the shallow CNN backbone layers.

Given $F_{ll}$, we insert random noise into the feature channel statistics using Normalization Perturbation (NP) as follows:
\begin{align}
& \hat{F_{ll}} = \mathrm{NP}(F_{ll}).
\end{align}
The Normalization Perturbation (NP) is formulated as:
\begin{align}
& y = \sigma_s^* \frac{x - \mu_c}{\sigma_c} + \mu_s^*, \\
& \sigma_s^* = \alpha \sigma_c, \\
& \mu_s^* = \beta \mu_c,
\end{align}
where $\{\mu_c, \sigma_c\} \in \mathbb{R}^{C} \text{ and } \{\mu_s, \sigma_s\} \in \mathbb{R}^{C}$ represent the mean and variance of the input content image and the stylized image, respectively.

Afterward, $\hat{F_{ll}}$ and $F_{high}$ are fed into the inverse discrete wavelet transform (IDWT) layer to reconstruct the style-augmented feature map $\hat{F} \in \mathbb{R}^{H \times W \times C}$. The overall process is illustrated in Fig. \ref{fig2:wavenp}. In this letter, we refer to our style augmentation module as WaveNP. Following~\cite{fan2023towards}, WaveNP is applied at stages 1 and 2 of the CNN backbone during model training.

\subsection{Domain-agnostic Query Selection}
In DETR models, object queries are a fixed set of learnable embeddings that serve as input to the Transformer decoder. These queries serve as latent object representations that interact with global image features in the decoder and are progressively optimized to map to object instances in images. Since optimizing learnable object queries is inherently challenging, several methods~\cite{zhudeformable, zhangdino} have introduced query selection schemes that use confidence scores to select the top-K features as initial object queries. Domain shifts, which primarily comes from variations in visual style, introduce significant style-induced biases (\textit{i.e.}, domain-induced biases) into latent features, thereby limiting the generalization ability of the learned model. Therefore, removing domain-induced biases from queries enhances the robustness of DETR models against real-world domain shifts.

\begin{figure}[t]
 \centering
    \includegraphics[scale=0.38]{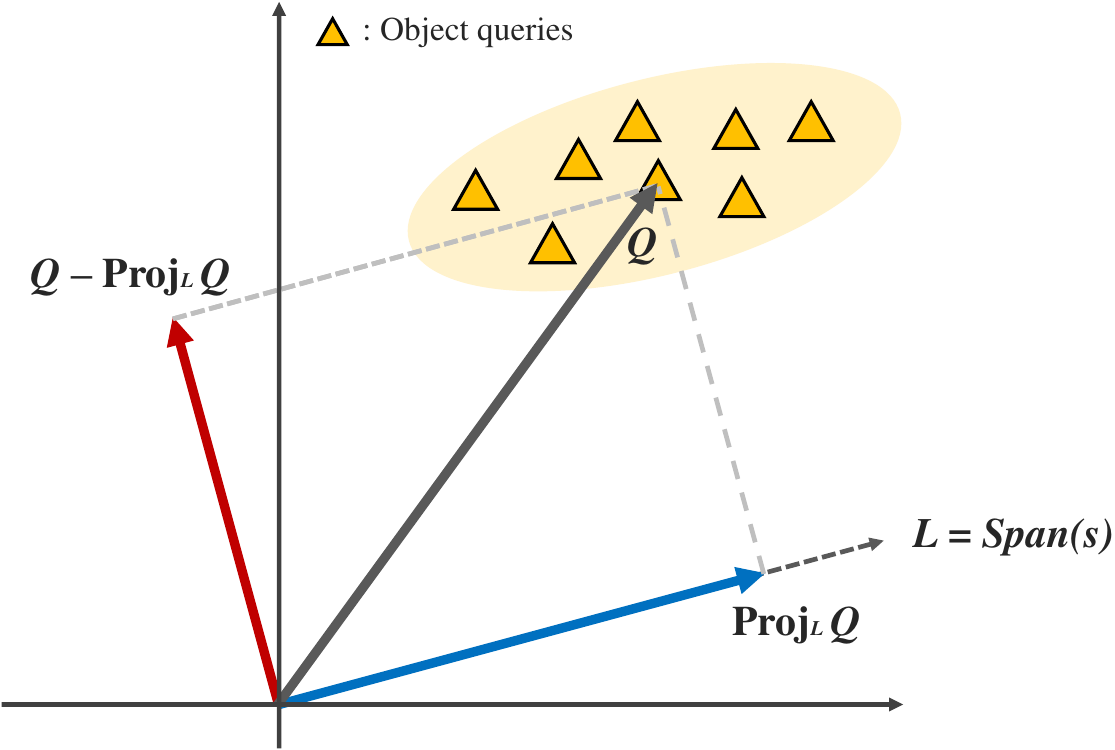}
    \caption{Visualization of the orthogonal projection. The query $Q$ is projected onto style-irrelevant semantic axes.}
    \label{fig3:daqs}
    \vspace{0.2cm}
 \end{figure}

To address this problem, we propose a Domain-agnostic Query Selection which removes style-induced biases from object latent representations. As illustrated in Fig. \ref{fig3:daqs}, our method leverages the orthogonality in the latent space to project queries onto style-irrelevant semantic axes. Given style-related statistics (\textit{i.e.}, $\mu_s \text{~and~} \sigma_s$), we construct a latent style representation as follows:
 \begin{align}
& s = E_s(\mu_s + \sigma_s),
\end{align}
where $s \in \mathbb{R}^{D}$ is a style embedding, $D$ is the feature dimension, and $E_{s}$ is a style encoder consisting of a linear layer and a normalization layer.
                    
Let $L$ represent the subspace spanned by $s$ and $Q \in \mathbb{R}^{N \times D}$, where $N$ is the number of feature sequences, denote the flattened encoded image features. We aim to remove the component of $Q$ that lies along the axis of $L$:
\begin{align}
\hat{Q} = Q - \alpha \text{Proj}_{L} Q,
\label{eq6}
\end{align}
where $\alpha$ is a hyperparameter between 0 and 1 that controls the degree of removal of the style components. Note that during training, $\alpha$ is fixed at 1.

From $\hat{Q}$, we use confidence scores to select the top-K scoring features as the initial object queries:
 \begin{align}
& \hat{Q}_{select} = \text{Top-}K(E_c(\hat{Q})),
\end{align}
where $\hat{Q}_{select}$ 
represents the set of K selected features, and $E_c$ is an auxiliary prediction head to select the top-K features.

We note that removing style-induced biases from image features may incur the loss of meaningful information. Therefore we subtract style-relevant components from feature sequences only in query selection process.
\vspace{0.2cm}

\begin{table*}[t]
    \centering
    \caption{Quantitative comparison of domain generalization performance on the Diverse Weather Dataset (DWD). The results represent mAP (\%) for different weather conditions when trained on Daytime-Sunny and tested on Night-Sunny, Dusk-Rainy, Night-Rainy, and Daytime-Foggy. Bold numbers indicate the best performance, while underlined numbers denote the second-best performance.
    }
    \scriptsize
    \resizebox{\textwidth}{!}{
    
    \begin{tabular}{l | l | c | ccccccc >{\columncolor{gray!30}}c | ccccccc >{\columncolor{gray!30}}c}
        \toprule
        \multirow{2}{*}{} & \multirow{2}{*}{} & \multirow{2}{*}{} 
        & \multicolumn{8}{c|}{\textbf{Night-Sunny}} 
        & \multicolumn{8}{c}{\textbf{Dusk-Rainy}} \\
        \cmidrule(lr){4-11} \cmidrule(lr){12-19}
        Method & Backbone & Detector & bus & bike & car & motor & person & rider & truck & mAP 
        & bus & bike & car & motor & person & rider & truck & mAP \\
        \midrule
        Faster R-CNN~\cite{fasterrcnn} & ResNet-101 & -
        & 37.7 & 30.6 & 49.5 & 15.4 & 31.5 & 28.6 & 40.8 & 33.5
        & 36.8 & 15.8 & 50.1 & 12.8 & 18.9 & 12.4 & 39.5 & 26.6 \\
        RT-DETR~\cite{zhao2024detrs} & ResNet-50 & - 
        & 45.4 & \underline{41.2} & 70.5 & 19.1 & 48.9 & 30.8 & 47.6 & 43.3
        & 46.1 & 23.7 & 74.0 & 16.1 & 36.7 & 16.3 & 49.7 & 37.5 \\
        CDSD~\cite{wu2022single} & ResNet-101 & Faster-RCNN
        & 40.6 & 35.1 & 50.7 & 19.7 & 34.7 & 32.1 & 43.4 & 36.6
        & 37.1 & 19.6 & 50.9 & 13.4 & 19.7 & 16.3 & 40.7 & 28.2 \\
        CLIP-Gap~\cite{vidit2023clip} & ResNet-101 & Faster-RCNN
        & 37.7 & 34.3 & 58.0 & 19.2 & 37.6 & 28.5 & 42.9 & 36.9
        & 37.8 & 22.8 & 60.7 & 16.8 & 26.8 & 18.7 & 42.4 & 32.3 \\
        SW~\cite{pan2019switchable} & ResNet-50 & RT-DETR 
        & 34.6 & 23.7 & 65.7 & 10.2 & 37.1 & 19.7 & 35.2 & 32.3
        & 40.0 & 14.2 & 72.6 &  6.8 & 34.7 & 10.5 & 41.6 & 31.5 \\
        IBN-Net~\cite{pan2018two} & ResNet-50 & RT-DETR 
        & 36.8 & 24.0 & 65.5 & 13.1 & 39.9 & 21.2 & 38.9 & 34.2
        & 43.2 & 16.0 & 72.3 & 14.4 & 35.4 & 12.2 & 44.9 & 34.1 \\
        IterNorm~\cite{huang2019iterative} & ResNet-50 & RT-DETR 
        & 36.8 & 31.9 & 68.4 & 14.6 & 44.9 & 25.7 & 38.8 & 37.3
        & 35.4 & 17.4 & 69.1 & 15.7 & 29.7 & 11.9 & 39.7 & 31.3 \\
        NP~\cite{fan2023towards} & ResNet-50 & RT-DETR 
        & \underline{49.0} & 40.5 & \underline{71.9} & \underline{20.8} & \underline{49.2} & \underline{33.4} & \underline{50.2} & \underline{45.0}
        & \textbf{51.8} & \textbf{28.6} & \underline{77.4} & \underline{19.0} & \textbf{42.2} & \textbf{22.3} & \underline{54.1} & \textbf{42.2} \\
        \midrule
        \textbf{DG-DETR (ours)} & ResNet-50 & RT-DETR 
        & \textbf{51.2} & \textbf{44.7} & \textbf{73.4} & \textbf{23.0} & \textbf{51.9} & \textbf{36.1} & \textbf{52.5} & \textbf{47.6}
        & \underline{51.0} & \underline{28.1} & \textbf{77.7} & \textbf{20.1} & \underline{42.0} & \underline{19.9} & \textbf{55.7} & \underline{42.1} \\
        \bottomrule
    \end{tabular}}
    
    \vspace{1em}
    
    \centering
    \scriptsize
    \resizebox{\textwidth}{!}{
    \begin{tabular}{l | l | c | ccccccc >{\columncolor{gray!30}}c | ccccccc >{\columncolor{gray!30}}c}
        \toprule
        \multirow{2}{*}{} & \multirow{2}{*}{} & \multirow{2}{*}{} 
        & \multicolumn{8}{c|}{\textbf{Night-Rainy}} 
        & \multicolumn{8}{c}{\textbf{Daytime-Foggy}} \\
        \cmidrule(lr){4-11} \cmidrule(lr){12-19}
        Method & Backbone & Detector & bus & bike & car & motor & person & rider & truck & mAP 
        & bus & bike & car & motor & person & rider & truck & mAP \\
        \midrule
        Faster R-CNN~\cite{fasterrcnn} & ResNet-101 & -
        & 22.6 & 11.5 & 27.7 & 0.4 & 10.0 & 10.5 & 19.0 & 14.5
        & 30.7 & 26.7 & 49.7 & 26.2 & 30.9 & 35.5 & 23.2 & 31.9 \\
        RT-DETR~\cite{zhao2024detrs} & ResNet-50 & - 
        & 27.7 &  7.8 & 42.0 &  0.3 & 12.9 &  8.7 & 24.8 & 17.7
        & 32.7 & 27.5 & \underline{61.8} & 30.6 & 38.1 & 36.9 & 25.0 & 36.1 \\
        CDSD~\cite{wu2022single} & ResNet-101 & Faster-RCNN
        & 24.4 & 11.6 & 29.5 & \textbf{9.8} & 10.5 & \underline{11.4} & 19.2 & 16.6
        & 32.9 & 28.0 & 48.8 & 29.8 & 32.5 & 38.2 & 24.1 & 33.5 \\
        CLIP-Gap~\cite{vidit2023clip} & ResNet-101 & Faster-RCNN
        & 28.6 & 12.1 & 36.1 & \underline{9.2} & 12.3 & 9.6 & 22.9 & 18.7
        & \underline{36.1} & \textbf{34.3} & 58.0 & \textbf{33.1} & \underline{39.0} & \textbf{43.9} & 25.1 & \textbf{38.5} \\
        SW~\cite{pan2019switchable} & ResNet-50 & RT-DETR 
        & 20.3 &  6.5 & 42.9 &  0.1 & 10.8 &  4.1 & 18.7 & 14.8
        & 23.5 & 21.4 & 55.7 & 20.4 & 31.1 & 32.7 & 12.9 & 28.2 \\
        IBN-Net~\cite{pan2018two} & ResNet-50 & RT-DETR 
        & 29.3 &  6.6 & 42.6 &  0.9 & 13.7 &  8.0 & 24.9 & 18.0
        & 19.9 & 17.8 & 47.2 & 19.3 & 25.7 & 24.1 & 13.0 & 23.9 \\
        IterNorm~\cite{huang2019iterative} & ResNet-50 & RT-DETR 
        & 18.6 &  7.4 & 33.6 &  0.6 &  9.1 &  7.9 & 12.8 & 12.9
        & 25.9 & 25.5 & 59.2 & 24.7 & 37.4 & 36.1 & 15.6 & 36.2 \\
        NP~\cite{fan2023towards} & ResNet-50 & RT-DETR 
        & \textbf{39.9} & \underline{16.8} & \underline{50.4} &  1.0 & \underline{17.7} &  8.9 & \underline{32.0} & \underline{23.9}
        & 33.2 & 27.1 & 61.7 & 29.2 & 38.3 & 37.6 & \underline{25.9} & 36.2 \\
        \midrule
        \textbf{DG-DETR (ours)} & ResNet-50 & RT-DETR 
        & \underline{39.1} & \textbf{19.6} & \textbf{51.3} & 4.5 & \textbf{18.3} & \textbf{14.0} & \textbf{32.7} & \textbf{25.6}
        & \textbf{36.4} & \underline{28.9} & \textbf{64.4} & \underline{31.2} & \textbf{40.3} & \underline{39.9} & \textbf{28.8} & \textbf{38.5} \\
        \bottomrule
    \end{tabular}}
    \label{tab1:dwd}
    \vspace{0.2cm}
\end{table*}

\begin{table*}[t]
    \centering
    \caption{The performance comparison with state-of-the-art methods on Cityscapes-C. 
    The Avg. means the average performance across 15 corruption types, each evaluated at 5 different severity levels. Bold numbers indicate the best performance, while underlined numbers denote the second-best performance.
    }
    \scriptsize
    \resizebox{\textwidth}{!}{
    \begin{tabular}{l|c|ccc|cccc|cccc|cccc|>{\columncolor{gray!30}}c}
        \toprule
        & & \multicolumn{3}{c|}{Noise} & \multicolumn{4}{c|}{Blur} & \multicolumn{4}{c|}{Weather} & \multicolumn{4}{c|}{Digital} & \\
        \cmidrule(lr){3-5} \cmidrule(lr){6-9} \cmidrule(lr){10-13} \cmidrule(lr){14-17}
        Method & Clean & Gauss & Shot & Impulse & Defocus & Glass & Motion & Zoom & Snow & Frost & Fog & Bright & Contrast & Elastic & Pixel & JPEG & Avg. \\
        \midrule
        SupCon~\cite{khosla2020supervised} & 43.2 & 7.0 & 9.5 & 7.4 & 22.6 & 20.2 & 22.3 & 4.3 & 5.3 & 23.0 & 37.3 & 38.9 & \underline{31.6} & 40.1 & 24.0 & 20.1 & 20.9 \\
        FSCE~\cite{sun2021fsce} & 43.1 & 7.4 & 10.2 & 8.2 & 23.3 & 20.3 & 21.5 & \underline{4.8} & 5.6 & 23.6 & 37.1 & 38.0 & \underline{31.9} & 40.0 & 23.2 & 20.4 & 21.0 \\
        OA-DG~\cite{lee2024object} & 43.4 & 8.2 & 10.6 & 8.4 & 24.6 & 20.5 & 22.3 & \underline{4.8} & 6.1 & \underline{25.0} & \underline{38.4} & 39.7 & \textbf{32.8} & 40.2 & 23.8 & 22.0 & 21.8 \\
        PhysAug~\cite{xu2025physaug} & 42.6 & 14.3 & 17.0 & 11.9 & 25.6 & 19.1 & 25.5 & 3.9 & 8.6 & 21.3 & 35.3 & 39.5 & 27.5 & 39.1 & 28.9 & 19.9 & 22.6 \\
        \midrule
        RT-DETR~\cite{zhao2024detrs} & \underline{50.4} & \underline{21.9} & \underline{26.0} & \underline{19.2} & \underline{31.6} & \underline{33.7} & \underline{31.0} & 3.7 & \underline{14.4} & 20.6 & 38.1 & \underline{47.2} & 29.5 & \underline{47.0} & \underline{47.1} & \underline{39.8} & \underline{30.1}  \\
        
        \textbf{DG-DETR (ours)} & \textbf{52.2} & \textbf{28.2} & \textbf{33.2} & \textbf{26.4} & \textbf{37.0} & \textbf{40.6} & \textbf{36.0} & \textbf{5.5} & \textbf{24.5} & \textbf{29.2} & \textbf{43.1} & \textbf{50.9} & 30.9 & \textbf{50.4} & \textbf{50.1} & \textbf{43.1} & \textbf{35.3} \\
        \bottomrule
    \end{tabular}
    }
    \label{tab:cityscapes-c}
    \vspace{0.2cm}
\end{table*}

\begin{table}[h]
\centering
\caption{In-domain performance comparison on the Daytime-Sunny of the Diverse Weather Dataset (DWD). All models are trained and tested under the same weather condition. Bold numbers represent the best performance, while underlined numbers denote the second-best performance.
}
\scriptsize
\setlength{\tabcolsep}{3pt} 
\begin{tabular}{l|ccccccc|>{\columncolor{gray!30}}c}
    \toprule
    \multirow{2}{*}{} & \multicolumn{8}{c}{\textbf{Daytime-Sunny}} \\
    \cmidrule(lr){2-9}
    Method & bus & bike & car & motor & person & rider & truck & mAP \\
    \midrule
    Faster R-CNN~\cite{fasterrcnn} & \underline{63.4} & 42.9 & 53.4 & 49.4 & 39.8 & 48.1 & 60.8 & 51.1 \\
    RT-DETR \cite{zhao2024detrs} & 61.3 & 48.3 & 82.2 & 51.7 & 59.8 & 49.2 & 63.1 & 59.4 \\
    CDSD~\cite{wu2022single} & \textbf{68.8} & \textbf{50.9} & 53.9 & \textbf{56.2} & 41.8 & \underline{52.4} & \textbf{68.7} & 56.1 \\
    CLIP-Gap~\cite{vidit2023clip} & 55.0 & 47.8 & 67.5 & 46.7 & 49.4 & 46.8 & 54.7 & 52.6 \\
    SW \cite{pan2019switchable} & 55.4 & 40.4 & 80.5 & 42.9 & 54.5 & 44.5 & 55.9 & 53.4 \\
    IBN-Net \cite{pan2018two} & 56.0 & 37.3 & 80.0 & 42.7 & 53.9 & 43.1 & 57.3 & 52.9 \\
    IterNorm \cite{huang2019iterative} & 60.4 & 47.3 & 82.4 & 46.2 & 59.7 & 47.9 & 61.2 & 57.9 \\
    NP \cite{fan2023towards} & 61.1 & 49.7 & \underline{82.9} & 49.6 & \underline{60.5} & 51.9 & 63.4 & \underline{59.9} \\
    \midrule
    \textbf{DG-DETR (ours)} & 62.5 & \underline{50.7} & \textbf{83.5} & \underline{51.9} & \textbf{61.6} & \textbf{53.3} & \underline{64.9} & \textbf{61.2} \\
    \bottomrule
\end{tabular}
\label{tab2:source}
\vspace{0.2cm}
\end{table}

\begin{table}[h]
    \centering
    \caption{Ablation analysis of the proposed components. The results presents the impact of Wavelet-guided Style Augmentation (WaveNP) and Domain-agnostic Query Selection (DAQS) on domain generalization performance. DS, NS, DR, NR, and DF stand for daytime-sunny, night-sunny, dusk-rainy, night-rainy, and daytime-foggy, respectively.
    }
    \scriptsize
    \begin{tabular}{cc|c|cccc}
        \toprule
        \textbf{WaveNP} & \textbf{DAQS} & \textbf{DS} & \textbf{NS} & \textbf{DR} & \textbf{NR} & \textbf{DF} \\
        \midrule
        \textcolor{lightgray}{\checkmark} & \textcolor{lightgray}{\checkmark} & 59.4 & 43.3 & 37.5 & 17.7 & 36.1 \\
        \checkmark & \textcolor{lightgray}{\checkmark} & 60.7 & 46.7 & \textbf{42.4} & 25.0 & 37.7 \\
        \textcolor{lightgray}{\checkmark} & \checkmark & 60.6 & 46.3 & 39.2 & 20.9 & 38.4 \\
        \checkmark & \checkmark & \textbf{61.2} & \textbf{47.6} & 42.1 & \textbf{25.6} & \textbf{38.5} \\
        \bottomrule
    \end{tabular}
    \label{tab3:abl_module}
\vspace{0.2cm}
\end{table}

\begin{table}[h]
    \centering
    \caption{Ablation analysis on perturbation of different frequency components. We compares the impact of applying feature perturbation to high-frequency vs. low-frequency components on domain generalization performance. 
    }
    \scriptsize
    \begin{tabular}{cc|c|cccc}
        \toprule
        \textbf{High} & \textbf{Low} & \textbf{DS} & \textbf{NS} & \textbf{DR} & \textbf{NR} & \textbf{DF} \\
        \midrule
        \checkmark & \textcolor{lightgray}{\checkmark} & 59.3 & 43.8 & 37.8 & 18.5 & 36.2 \\
        \textcolor{lightgray}{\checkmark} & \checkmark & \textbf{60.7} & \textbf{46.7} & \textbf{42.4} & \textbf{25.0} & \textbf{37.7} \\
        \bottomrule
    \end{tabular}
    \label{tab4:abl_frq}
    \vspace{0.2cm}
\end{table}

\begin{table}[h]
    \centering
    \caption{Ablation analysis of DG-DETR and its compatibility with different DETR-based detectors. The results show the performance of DG-DETR components integrated with RT-DETR and DINO.
    }
    \scriptsize
    \scalebox{0.95}{
        \begin{tabular}{llc|ccccc}
            \toprule
            \textbf{Method} & \textbf{Backbone} & \textbf{Detector} & \textbf{DS} & \textbf{NS} & \textbf{DR} & \textbf{NR} & \textbf{DF} \\
            \midrule
            RT-DETR~\cite{zhao2024detrs} & ResNet-50 & - & 59.4 & 43.3 & 37.5 & 17.7 & 36.1 \\
            DG-DETR (ours) & ResNet-50 & RT-DETR & \textbf{61.2} & \textbf{47.6} & \textbf{42.1} & \textbf{25.6} & \textbf{38.5} \\
            \midrule
            DINO~\cite{zhangdino} & ResNet-50 & - & 57.5 & 36.1 & 29.3 & 10.5 & 34.6 \\
            DG-DETR (ours) & ResNet-50 & DINO & \textbf{58.7} & \textbf{38.8} & \textbf{31.4} & \textbf{13.0} & \textbf{35.9} \\
            \bottomrule
        \end{tabular}
    }
    \label{tab:compatibility_ablation}
    \vspace{0.2cm}
\end{table}

\section{Experiments}

In this section, we evaluate the generalization capability of our method against out-of-distribution scenarios including an ablation study to verify the effectiveness of proposed components.

\subsection{Experimental setup}
To evaluate our method, we use the Diverse Weather Dataset (DWD)~\cite{wu2022single} and the Cityscapes-C~\cite{michaelis2019benchmarking}. DWD is an urban-scene detection benchmark that includes five different weather conditions: Daytime-Sunny (DS), Night-Sunny (NS), Night-Rainy (NR), Dusk-Rainy (DR), and Daytime-Foggy (DF). It was collected from BDD-100k~\cite{yu2020bdd100k}, FoggyCityscapes~\cite{sakaridis2018semantic}, and AdverseWeather~\cite{hassaballah2020vehicle}. Following~\cite{wu2022single}, we train the model using only the source domain (\textit{i.e.}, Daytime-Sunny) and evaluate it directly on the other adverse weather domains. Cityscapes-C is a robust detection benchmark built upon the Cityscapes dataset~\cite{cordts2016cityscapes} and provides 15 corruption types across 5 severity levels. We use the training set of Cityscapes as the source domain, and consider the corrupted versions of the validation set as unseen target domains.

In experiments, we adopt RT-DETR~\cite{zhao2024detrs} as the base detector. To the best of our knowledge, no prior study has explored domain generalization (DG) in DETR-based object detection. Therefore, we extend existing DG methods~\cite{pan2019switchable, pan2018two, huang2019iterative, choi2021robustnet, fan2023towards}, originally implemented for CNNs, to RT-DETR. All these methods improve the generalization ability of models by applying feature normalization ~\cite{pan2019switchable, pan2018two, huang2019iterative, choi2021robustnet} or perturbation~\cite{fan2023towards} to the CNN feature extractor. Thus, we directly integrate them into DETR's feature extractor (\textit{i.e.}, CNN backbone). To evaluate domain generalization performance, we follow~\cite{wu2022single} and use the Mean Average Precision (mAP) metric. Specifically, we report mAP at an Intersection over Union (IoU) threshold of 0.5 (mAP@0.5). The projection scaling factor (Eq. \ref{eq6}) is set to 1.0.

 \begin{figure*}[ht]
 \centering
    \includegraphics[scale=0.50]{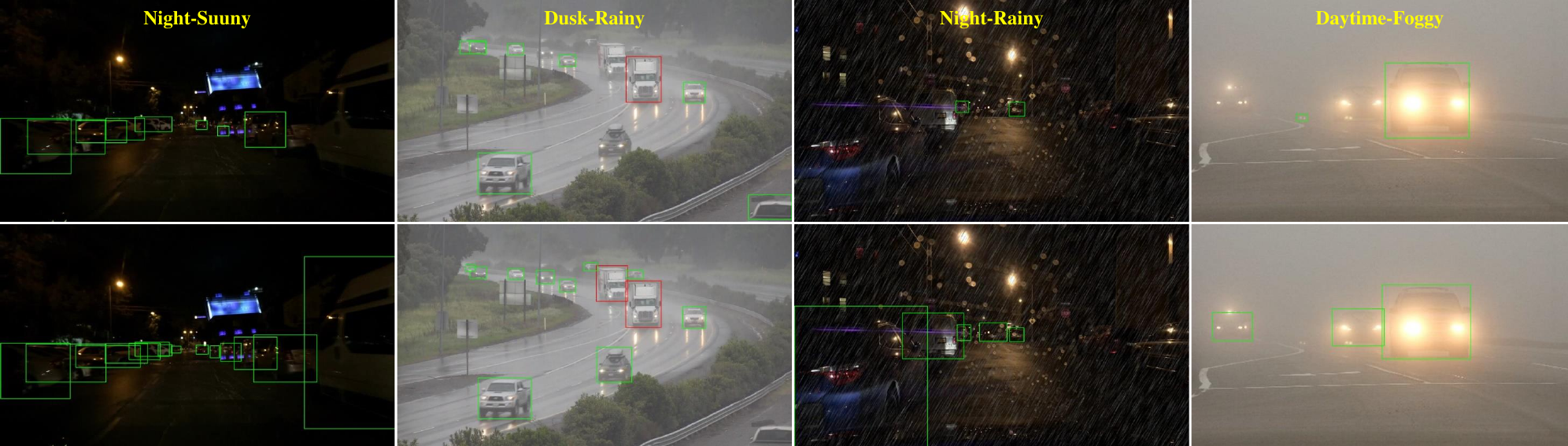}
    \caption{Detection results on Night-Sunny, Dusk-Rainy, Night-Rainy, Daytime-Foggy. \textbf{First Row:} RT-DETR \cite{zhao2024detrs} results. \textbf{Second Row:} The results with our method. Green bounding boxes indicate cars, and red bounding boxes indicate trucks.
    }
    \label{fig4:viscom}
    \vspace{0.2cm}
 \end{figure*}

\subsection{Performance Analysis of Domain Generalization}
Table~\ref{tab1:dwd} presents the domain generalization (DG) performance under real-world weather conditions, each representing different domain shift scenarios. Our method achieves the best performance on three of the datasets. Specifically, compared with the baseline, DG-DETR improves performance by 4.3\%, 4.6\%, 7.9\%, and 2.4\% on Night-Sunny (NS), Dusk-Rainy (DR), Night-Rainy (NR), and Daytime-Foggy (DF), respectively. On the other hand, normalization-based methods\cite{pan2019switchable, pan2018two, huang2019iterative, choi2021robustnet} fail to improve out-of-distribution (OOD) robustness. This is because these methods negatively affect the feature discrimination ability, thereby weakening detection performance, as reported in\cite{wu2022single}.

To further validate the robustness of our method, we evaluate its performance on the Cityscapes-C dataset, a benchmark for common corruptions. As shown in Table~\ref{tab:cityscapes-c}, DG-DETR significantly outperforms both the baseline RT-DETR and existing state-of-the-art methods. Our method achieves an average mAP of 35.3\%, a substantial improvement over the baseline (30.1\%) and the previous best method (22.6\%). These results indicate that DG-DETR effectively generalizes to both adverse weather conditions and diverse common corruptions.

We also conducted comparative experiments in in-domain scenarios, and the results are reported in Table~\ref{tab2:source}. The results show that DG-DETR significantly outperforms existing methods, demonstrating that the proposed approach improves performance even when the training and test sets come from the same domain. A similar trend is observed in the Cityscapes-C dataset, where DG-DETR achieves a performance of 52.2\% on the clean (source domain) domain, surpassing the baseline RT-DETR's 50.4\%.

\subsection{Ablation Analysis}
To further evaluate the effectiveness of our key modules and strategies in DG-DETR, we conduct a series of ablation experiments. Table~\ref{tab3:abl_module} presents the ablation study on each component. As observed, removing any single component degrades the DG performance, which demonstrates the contribution of each module. We note that Domain-agnostic Query Selection (DAQS) improves model performance even without augmentation or normalization techniques by removing domain-induced biases from object queries.

Table~\ref{tab4:abl_frq} presents an ablation analysis on the effect of feature perturbation to different frequency components. Perturbing only high-frequency components results in lower performance across all weather conditions. In contrast, perturbing low-frequency components leads to performance improvement, demonstrating that low-frequency perturbation effectively improves generalization ability by learning domain-invariant representations.

We further evaluate the plug-and-play capability of DG-DETR by integrating it with different DETR-based detectors~\cite{zhao2024detrs, zhangdino}. As shown in Table~\ref{tab:compatibility_ablation}, our method consistently improves the generalization performance of both RT-DETR and DINO, which highlights its compatibility and effectiveness across different frameworks.
\vspace{0.2cm}

\subsection{Qualitative Results}
We present qualitative comparisons across different weather scenarios in Fig.~\ref{fig4:viscom}. As observed, the baseline RT-DETR fails to detect objects in challenging environments, often missing detections or producing inaccurate predictions. In contrast, our method shows more precise and reliable object detection under adverse conditions, demonstrating its robustness in cross-domain scenarios.
\vspace{0.2cm}

\section{Conclusion}
In this letter, we propose Domain Generalized DEtection TRansformer (DG-DETR) for DETR-based single-domain generalization. WaveNP preserves object semantics during feature perturbation by disentangling domain-invariant and domain-specific features. Domain-agnostic query selection improves the OOD robustness by removing domain-induced biases from object queries. Experimental results demonstrate that DG-DETR significantly improves the generalization ability of DETR models to unseen domains.
\vspace{0.2cm}

\bibliographystyle{elsarticle-num}
\bibliography{bibliography}

\end{document}